\documentclass[10pt,letterpaper]{IEEEtran}
\usepackage{graphicx}
\usepackage{graphics}
\usepackage{amssymb}
\usepackage{amsmath}
\usepackage{color}
\usepackage{xspace}
\usepackage{algpseudocode}
\usepackage{bbm}
\usepackage{comment}
\usepackage{algorithm}
\usepackage{algorithmicx}
\usepackage{subfig}
\usepackage{soul}
\usepackage{array}
\usepackage[bottom]{footmisc}
\usepackage{multirow}
\usepackage{cancel}
\usepackage{hyperref} 
\usepackage{xr}
\usepackage{cite}

\definecolor{CranJ}{cmyk}{0,0.69,0.54,0.04} 
\definecolor{PinkJ}{cmyk}{0,0.71,0.43,0.12} 
\definecolor{Cran}{cmyk}{0,0.73,0.41,0.29} 
\definecolor{VRed}{cmyk}{0,0.75,0.25,0.2} 
\definecolor{ORed}{cmyk}{0,0.75,0.75,0} 
\definecolor{CBlue}{cmyk}{1,0.25,0,0} 
\setstcolor{VRed}
\usepackage{bm}

\newlength\myindent
\makeatletter

\title{\LARGE \bf Cooperative Localization under Limited Connectivity
}

\author{Jianan Zhu \quad Solmaz S. Kia\\
\normalsize{\emph{University of California Irvine}}
  \thanks{The authors
    are with the Mechanical and Aerospace Eng. Dept. of 
    Univ. of California Irvine, CA 92697,~USA, {\tt\small jiananz1,solmaz@uci.edu}. This work is supported by the U.S. Dept. of Commerce, National Institute of Standards and Technology award 70NANB17H192.
    }%
}

\newcommand{\real}{{\mathbb{R}}}

\newcommand{\argmin}{\operatorname{argmin}}

\newcommand{\prpg}{\mbox{{\footnotesize\textbf{--}}}}
\newcommand{\updt}{\mbox{\textbf{+}}}

\newcommand{\until}[1]{\in\{1,\cdots,#1\}}

\newcommand{\vect}[1]{\boldsymbol{\mathbf{#1}}}
\newcommand{\Bvect}[1]{\bm\bar{\boldsymbol{\mathbf{#1}}}}

\newcommand{\Hvect}[1]{\bm\hat{\boldsymbol{\mathbf{#1}}}}

 \newcommand{\boxend}{\hfill \ensuremath{\Box}}
\newcommand{\oprocendsymbol}{\hbox{$\bullet$}}
\newcommand{\oprocend}{\relax\ifmmode\else\unskip\hfill\fi\oprocendsymbol}

 \newcommand{\blue}[1]{{\color{black} #1}}

\allowdisplaybreaks

\newtheorem{thm}{Theorem}[section]
\newtheorem{rem}{Remark}[section]

\newtheorem{lem}{Lemma}[section]

\begin{document}
\maketitle
\begin{abstract}
We report two decentralized multi-agent cooperative localization algorithms in which, to reduce the communication cost, inter-agent state estimate correlations are not maintained but accounted for implicitly. In our first algorithm, to guarantee filter consistency, we account for unknown inter-agent correlations via an upper bound on the joint covariance matrix of the agents. In the second method, we use an optimization framework to estimate the unknown inter-agent cross-covariance matrix. In our algorithms, each agent localizes itself in a global coordinate frame using a local filter driven by local dead reckoning and occasional absolute measurement updates, and opportunistically corrects its pose estimate whenever it can obtain relative measurements with respect to other mobile agents. To process any relative measurement, only the agent taken the measurement and the agent the measurement is taken from need to communicate with each other.
Consequently, our algorithms are decentralized algorithms that do not impose restrictive network-wide connectivity condition. Moreover, we make no assumptions about the type of agents or  relative measurements. We demonstrate our algorithms in simulation and a robotic~experiment.
\end{abstract}

\vspace{-0.06in}
\section{Introduction}
\vspace{-0.05in}
We consider the problem of decentralized cooperative localization (CL) for a group of mobile agents that cannot maintain persistent network connectivity. In CL  mobile agents
(e.g., mobile robots, human agents, unmanned vehicles) improve their local pose estimates using inter-agent relative measurement feedbacks. CL is often used in applications where access to external landmarks and aiding signals such as global positioning system (GPS) is challenging, see e.g.~\cite{POA-CR-RKM:01,AB-MRW-JJL:09,HL-FN:13,SEW-JMW-LLW-RME:13,JZ-SSK:18}.
Relative measurement updates in CL creates strong correlations among state estimates of the agents. 
Correlation between any two agents creates coupling terms in their estimation equations. 
Therefore, to maintain the correlation terms, agents need to communicate in a persistent manner at each timestep. The correlations cannot be ignored, as it will cause the so-called rumor propagation phenomenon that can lead to~overconfidence and, even to estimate divergence as reported~in~\cite{POA-CR-RKM:01}. 

Joint CL, which treats the team of mobile agents as one system and processes the inter-agent measurements to update the state estimate of all the agents, delivers the best localization accuracy. This is because the prior correlations allow agents other than the two involved in a relative measurement also benefit from relative measurement update (see~\cite{SSK-SF-SM:16} for further discussions). However, decentralized implementation of a joint CL in its naive form requires all-to-all or all-to-a-fusion-center communication at each timestep. To reduce the communication cost, \cite{SIR-GAB:02,SSK-SF-SM:16,SSK-JH-DG-SM:18} use decomposition techniques to fully decouple the propagation stage of a joint Extended Kalman Filter (EKF) based CL. However, at the update stage these algorithms require various forms of in-network connectivity. Moreover, 
~\cite{SIR-GAB:02} and~\cite{SSK-SF-SM:16} require each agent to have $O(N^2)$ processing~and storage capabilities, where $N$ is the size of the team. The algorithm in~\cite{SSK-JH-DG-SM:18} requires a server in the team. Other decentralized joint CL algorithms are studied in~\cite{EDN-SIR-AM:09} and \cite{KYKL-TDB-HHTL:10}. 
In some applications such as underwater vehicle localization, smart car localization, and first-responder human agent localization problems, maintaining multi-agent connectivity is challenging. Therefore, implementing decentralized joint CL algorithms 
may not be possible. The objective in this paper is to devise CL solutions that, to reduce communication cost, do not maintain the correlations but account for them in an implicit manner such that the consistency of the estimates are~preserved.

\emph{Literature review}: To relax network connectivity,~\cite{SEW-JMW-LLW-RME:13} proposes a leader-assistive CL
scheme for underwater vehicles. This method uses ranges and state information
from a single reference source (the server) with higher navigation
accuracy to improve localization  
accuracy  of underwater
vehicle(s) (the client(s)). There is no cooperation between the clients, and to benefit from CL, the clients need to stay in contact with the server. 
Alternatively, to relax connectivity,~\cite{AB-MRW-JJL:09,POA-CR-RKM:01,LCC-EDN-JLG-SIR:13,HL-FN:13,DM-NO-VC:13} do not maintain account of inter agent correlations. To provide consistent estimates,
\cite{AB-MRW-JJL:09} proposes an interleaved update algorithm in which only the agent taking the relative measurement updates its state. This method maintains a  bank of EKFs at each agent and, using an accurate bookkeeping of the identity of the agents involved in previous updates and the age of such information, updates each of these filters only  if the state of the filter is not correlated to the state of the landmark agent (the agent the relative measurement is taken
from). The main drawback is the growing size of information needed at each update time which increases the computational complexity of the algorithm. 
~\cite{POA-CR-RKM:01,LCC-EDN-JLG-SIR:13,HL-FN:13,DM-NO-VC:13} account~for the unknown inter-agent correlations 
 using Covariance Intersection
fusion (CIF) method.  The CIF fuses two or more tracks from same process when the correlations between tracks are unknown~\cite{SJJ-JKU:97,SJJ-JKU:01}. But, in CL, the local pose estimates of two different mobile agents are updated based on the feedback from a relative measurement between them. Thus, CIF-based CL techniques assume that each agent keeps a copy of the state estimate of the entire team locally. For example~\cite{POA-CR-RKM:01} uses such an approach for the localization of a group of space
vehicles communicating over a fixed ring topology. 
To avoid  keeping a copy of the state estimate of the entire team, ~\cite{LCC-EDN-JLG-SIR:13} proposes an algorithm in which an agent taking relative pose measurement uses this measurement and its current pose estimate to obtain and broadcast a pose
and the associated error covariance of its landmark agent. Then, the landmark agent uses the CIF
method to fuse the newly acquired
pose estimate with its own current estimate to increase its
estimation accuracy. Another example of the use of split CIF is given in~\cite{HL-FN:13} for intelligent transportation vehicles localization. \cite{DM-NO-VC:13} uses a common past-invariant ensemble Kalman pose estimation filter for intelligent vehicles.  This algorithm differs from the decentralized CIF methods only in use of ensembles in place of the means and covariances.  These technique crucially rely on relative pose measurements and cannot be applied for the more common cases of relative range and relative bearing measurements.  
Moreover, since CIF based methods use conservative bounds to account for missing cross-covariance information, these methods often deliver highly conservative~estimates.

\emph{Statement of contributions}:
We propose two methods to process relative measurements between two agents to improve their localization accuracy, when the past correlations between their state estimates is unknown. In the first method, we use an upper bound on the joint covariance matrix of the agents to account for the unknown inter-agent cross-covariance terms. This bound is reminiscent of the bound used in the CIF method, however, our method is different as it takes a direct approach to process relative measurement feedbacks, without requiring to reconstruct an state estimate from the relative pose measurement. Consequently, no assumption on the type of the inter-agent relative measurements is needed. Our second method trades in extra computation for a better localization performance while maintaining exactly the same inter-agent communication requirement. In this method, 
we aim to construct the unknown cross-covariance matrix using 
an optimization framework. We formally establish the consistency and performance guarantees of these two methods. We use our update methods to construct a CL algorithm in which each agent localizes itself in a global coordinate frame using a local filter driven by local dead reckoning and occasional absolute measurement updates, and opportunistically corrects its pose estimate whenever 
it can obtain relative measurements with respect to other mobile agents. To process any relative measurement, only the agent taken the measurement and the agent the measurement is taken from need to communicate with each other.
Consequently, our algorithm is a decentralized algorithm that does not impose any restrictive network-wide connectivity condition.  Moreover, we make no assumptions about the type of agents or relative measurements. Therefore, our algorithm can be used for heterogeneous multi-agent teams. Our CL method can be used as an add-on augmentation to improve self-localization accuracy of the mobile agents. That is, agents can implement any localization strategy such as dead-reckoning or GPS and when the accuracy via these methods is not satisfactory, they can seek assistance from other agents in their communication and relative measurement sensors' ranges without compromising the estimation consistency. Simulations and a robotic experiment demonstrate our results. 

\section{Problem definition and objective statement}\label{sec::prob_def}
In a team of 
mobile agents, each with sensing, computation and communication capabilities, let $\vect{x}^i\in\real^{n^i}$ be the local state of agent $i$ \blue{(the size of the team can change over time and is not necessarily known to the agents)}. The local state includes the global pose (position and orientation) states along with possibly other states that describe the equations of motion of the agent. 
The motion of agent $i$ is independent from others and is described by
$\vect{x}^i(t+1)=f(\vect{x}^i(t),\vect{u}^i(t)+\vect{\nu}^i_u)+\vect{\nu}^i_x$, where $\vect{u}^i$ can either be velocity command or self-motion measurement command obtained, e.g., from odometry or \blue{inertial measurement unit}. Here, $\vect{\nu}^i_u$ is the self-motion measurement noise (set to zero if control inputs are used) and $\vect{\nu}_x$ is the process noise. 
Each agent uses a local filter to obtain an estimate of its own state $\Hvect{x}^{i\prpg}(t)\in\real^{n^i}$ and its corresponding error covariance matrix $\vect{P}^{i\prpg}(t)\in\mathbb{S}^{++}_{n^i}$ at each time step $t\in\mathbb{Z}^+$ using its motion model and occasional access to absolute measurements through e.g. GPS or \blue{measurement from known landmarks}. Here, ${S}^{++}_{n}$ is the set of positive definite matrices of size $n$. We call $\text{bel}^{i\prpg}(t)=(\Hvect{x}^{i\prpg}(t),\vect{P}^{i\prpg}(t))$ the belief of agent $i$ at time $t$. 

\begin{figure}
    \centering
    \includegraphics[scale=0.3, trim=5pt 5pt 5pt 5pt,clip]{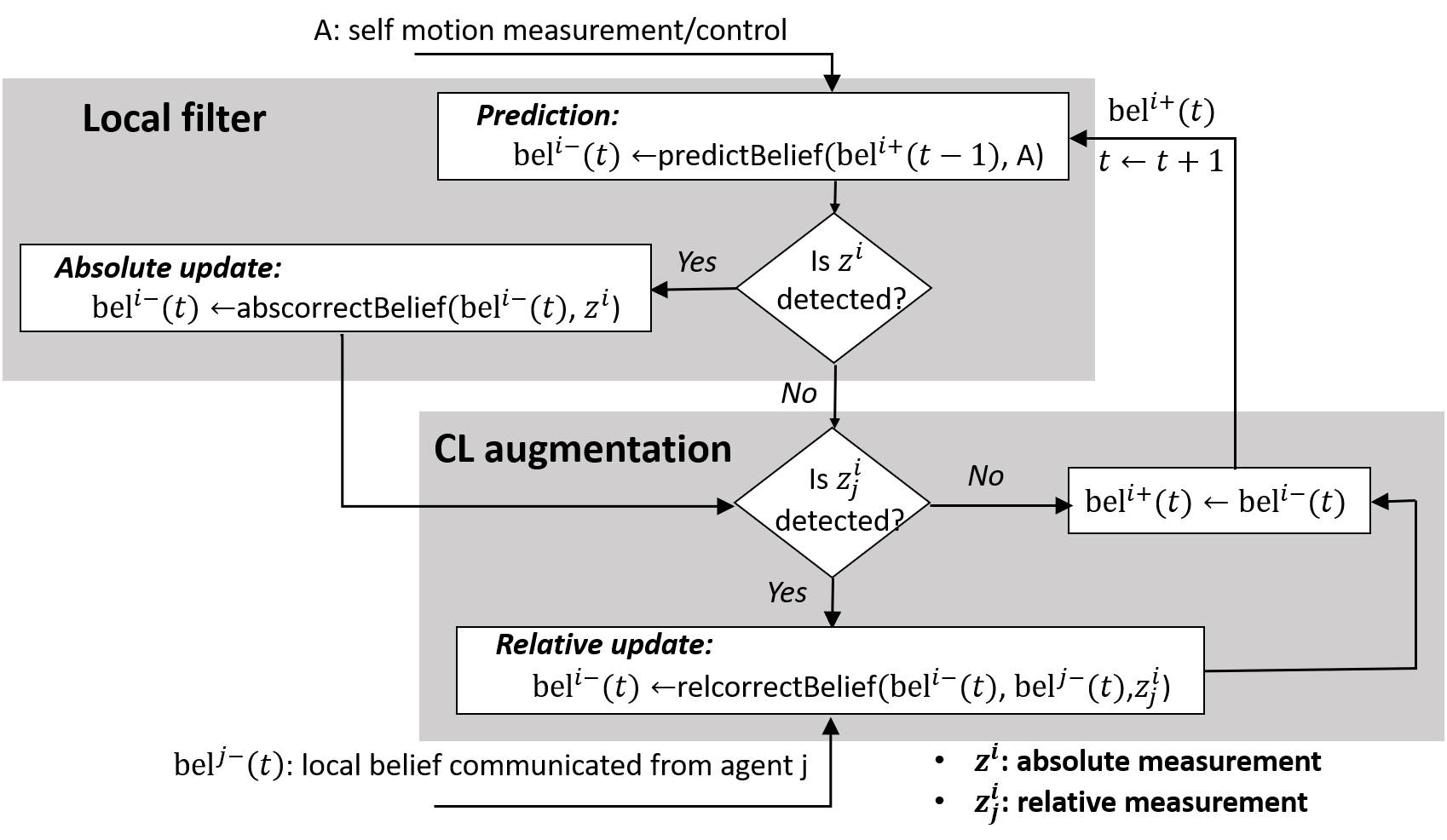}
    \caption{{\small CL as an augmentation atop of the local filter of agent $i$ becomes active when there is a relative measurement between agent $i$ and another agent $j$.}}\vspace{-0.15in}
    \label{fig::CL_diagram}
\end{figure}

Because of inherent noises in self-motion measurements and process noises, if access to absolute measurements is unreliable, the local filters will deliver poor estimates. To bound the error and improve accuracy, CL via joint processing of \blue{\emph{occasional}} available relative measurements among two agents is used.  Suppose that each agent has a set of exteroceptive sensors with limited sensing zone to detect, uniquely, the other agents in the team and to take relative measurements with respect to them, e.g., relative pose, relative range, and relative bearing or a combination of these measurements. \blue{We l}et the relative measurement taken by agent $i$ from agent $j$ at a time $t$ be denoted by $i\xrightarrow{t}j$ and described by
\begin{align}\label{eq::measur_ij}
  \vect{z}^i_{j}(t)&=\vect{h}^i_{j}(\vect{x}^i(t),\vect{x}^j(t))+\vect{\nu}^i(t),~~\vect{z}^i_{j}\in\real^{n_{z}^i}.
\end{align}
  We assume that all the sensor measurements are synchronized and also mutually independent. Moreover, $\vect{\nu}^i$ is white and zero mean Gaussian with $E[\vect{\nu}^i(t)\,\vect{\nu}^i(t)^\top]=\vect{R}^i>\vect{0}$, and $E[\vect{\nu}^i(k)\,\vect{\nu}^i(l)^\top]=\vect{0}$ for $k\neq l$. \blue{To relax network connectivity in CL, the ideal scenario is that any agent $i$ and agent $j$ should communicate with each other if and only if one of them has taken a relative measurement from another one, and at least one of them wants to process this relative measurement to improve its localization accuracy. This ideal operation is described in Fig.~\ref{fig::CL_diagram}, where functions $\mathrm{predictBelief}$ and $\mathrm{abscorrectBelief}$ denote the local localization filter of agent $i$, while function $\mathrm{relcorrectBelief}$ denotes the consistent relative measurement update method. Our objective is to design a relative measurement processing method that makes the CL in Fig.~\ref{fig::CL_diagram} possible. Next, we highlight the challenge of devising such a method.
  }

  \blue{
  According to the operation in Fig.~\ref{fig::CL_diagram}. if there is no relative measurement between an agent $i$ and other team members, the updated belief $\text{bel}^{i\updt}(t)=(\Hvect{x}^{i\updt}(t),\vect{P}^{i\updt}(t))$ of the agent is set to $\text{bel}^{i\prpg}(t)$ and is fed back to the local filter to produce the local belief at time $t+1$. On the other hand, if there is a relative measurement between agents $i$ and $j$ at time $t$, these agents can update their local belief as follows.} Let the joint state and the joint belief of the agents \blue{$\{i,j\}$} be, respectively, $\vect{x}_J(t)\!=\!(\vect{x}^{i}(t)^\top\!, \vect{x}^{j}(t)^\top)^\top$, and $\text{bel}_J^{\prpg}(t)\!=\!(\Hvect{x}_J^{\prpg}(t),\vect{P}_J^{\prpg}(t))$~where
\begin{align}\label{eq::P_J-actual}\Hvect{x}_J^{\prpg}(t)&=\begin{bmatrix}\Hvect{x}^{i\prpg}(t)\\ \Hvect{x}^{j\prpg}(t)\end{bmatrix},\quad
\vect{P}_J^{\prpg}(t)=\begin{bmatrix}\vect{P}^{i\prpg}(t)&\vect{P}_{ij}^{\prpg}(t)\\
{\vect{P}_{ij}^{\prpg}}^\top(t)&\vect{P}^{j\prpg}(t)\end{bmatrix}.\end{align}
\noindent\blue{Initially $\vect{P}^{\prpg}_{ij}(0)=\vect{0}$, but as shown below, $\vect{P}^{\prpg}_{ij}(t)$ is non-zero after 
a joint relative measurement update.} 
To simplify~the notation, hereafter we only include the time index $t$ when~clarification is needed. The state estimates are corrected according~to
\begin{align}\label{eq::our-x_c}
    \Hvect{x}^{l\updt}&=\Hvect{x}^{l\prpg}+\vect{K}^l\,(\vect{z}^i_{j}-\Hvect{z}^i_{j}),\quad l\in\{i,j\},
\end{align}
where $\Hvect{z}^i_{j}=\Hvect{z}_j^i=\vect{h}^i_{j}(\Hvect{x}^{i\prpg},\Hvect{x}^{j\prpg})$ is the estimated measurement. Let the first-order expansion of $\vect{h}_j^i(\vect{x}^{i},\vect{x}^{j})$ about the $\Hvect{x}_J^{\prpg}$ be
\begin{align}\label{eq::measur_ij}
  \vect{h}^i_{j}(\vect{x}^i,\vect{x}^j)\!&\!\approx
  \vect{h}^i_{j}(\Hvect{x}^{i\prpg},\Hvect{x}^{j\prpg})\!+\!\vect{H}^i_{i}\,(\vect{x}^i\!\!-\!\Hvect{x}^{i\prpg})\!+\!\vect{H}^i_{j}\,(\vect{x}^j\!\!-\!\Hvect{x}^{j\prpg}),
\end{align}
where $\vect{H}^i_{i}\!=\!\partial \vect{h}(\Hvect{x}^{i\prpg}\!,\Hvect{x}^{j\prpg})/\partial{\vect{x}}^i$ and $\vect{H}^i_{j}\!=\!\partial \vect{h}(\Hvect{x}^{i\prpg}\!,\Hvect{x}^{j\prpg})/\partial{\vect{x}}^j$.~Let
\begin{align}\label{eq::Jacobian_H}
\vect{H}^i&=\begin{bmatrix}
\vect{H}^i_{i}&\vect{H}_{j}^i
\end{bmatrix}.
\end{align}
\vspace{-0.1in}
\blue{\begin{lem}\label{lem::unbia}
If $E[\vect{\nu}^i]\!=\!\vect{0}$ and the prior belief of agent $l\in\{i,j\}$ is unbiased ($\text{E}[\vect{x}^{l}(t)-\Hvect{x}^{l\prpg}(t)]\!=\!\vect{0}$), the updated state~\eqref{eq::our-x_c} via any $\vect{K}^{l}$ is unbiased in the first-order approximate~sense. \boxend
\end{lem}
The proof of Lemma~\ref{lem::unbia} follows from standard results and is omitted for brevity. }

For $l\in\{i,j\}$, \blue{considering~\eqref{eq::our-x_c},} we have 
\begin{align}\label{eq::Ef_P_l}
&\mathrm{E}[(\vect{x}^l-\Hvect{x}^{l\updt})(\vect{x}^l-\Hvect{x}^{l\updt})^\top]\approx \mathrm{E}_f[(\vect{x}^l-\Hvect{x}^{l\updt})(\vect{x}^l-\Hvect{x}^{l\updt}\blue{)^\top}]=\nonumber\\
&\begin{bmatrix}(\vect{I}\!-\!\vect{K}^l\vect{H}_l^i)&-\vect{K}^l\vect{H}_k^i
\end{bmatrix}\begin{bmatrix}
 \vect{P}^{l\prpg}&\vect{P}_{lk}^{\prpg}\\
 {\vect{P}_{lk}^{\prpg}}^\top&\vect{P}^{k\prpg}
 \end{bmatrix}
\times\nonumber\\
&\begin{bmatrix}(\vect{I}\!-\!\vect{K}^l\vect{H}_l^i)&-\vect{K}^l\vect{H}_k^i
\end{bmatrix}^\top
\!\!+\vect{K}^l\vect{R}^i{\vect{K}^l}^\top\!\!,\quad k\in\{i,j\}\backslash\{l\}.\end{align}
We use $\mathrm{E}_f[.]$ to indicate that the expectation is taken over first-order approximate relative measurement model. After the update~\eqref{eq::our-x_c} with any \blue{non-zero} gain $\vect{K}^l$, \blue{even if $\vect{P}_{ij}^{\prpg}=\vect{0}$},  the state estimates are correlated \blue{because}  $
\mathrm{E}[(\vect{x}^i-\Hvect{x}^{i\updt})(\vect{x}^j-\Hvect{x}^{j\updt})^\top]\approx \mathrm{E}_f[(\vect{x}^i-\Hvect{x}^{i\updt})(\vect{x}^j-\Hvect{x}^{j\updt})^\top]=-(\vect{I}-\vect{K}^i\vect{H}_i^i)\vect{P}^{i\prpg}(\vect{K}^j\vect{H}_i^i)^\top-
(\vect{K}^i\vect{H}_j^i)\vect{P}^{j\prpg}(\vect{I}-\vect{K}^j\vect{H}_j^i)^\top+(\vect{I}-\vect{K}^i\vect{H}_i^i)\vect{P}^{\prpg}_{ij}(\vect{I}\!-\!\vect{K}^j\vect{H}_j^i)^\top+(\vect{K}^i\vect{H}_j^i){\vect{P}_{ij}^{\prpg}}^\top(\vect{K}^j\vect{H}_i^i)
\!+\vect{K}^i\vect{R}^i{\vect{K}^j}^\top\!\!$.

\textbf{Explicit minimum variance (EMV) relative measurement update}: 
 when $\vect{P}^{\prpg}_{ij}$ is known explicitly, the gain that gives a (sub-optimal) minimum variance estimate, similar to the EKF gain, can be obtained from
\begin{align}\label{eq::K-opt-indiv}
    \vect{K}_{\blue{\mathrm{EMV}}}^l\!=\!\underset{\blue{\vect{K}^l}}{\argmin} \text{Tr}(\mathrm{E}_f[(\vect{x}^l\!-\!\Hvect{x}^{l\updt})(\vect{x}^l\!-\!\Hvect{x}^{l\updt}]),~ l\!\in\!\{i,j\}.
\end{align}
Suboptimality is due to the linearization error in~\eqref{eq::measur_ij}. Let $\bm\tilde{\vect{H}}^{i,l}=\begin{bmatrix}\vect{H}^i_l&\vect{H}^i_k \end{bmatrix}$, then the solution of~\eqref{eq::K-opt-indiv} is (see~\cite{JLC-JLJ:12})
\vspace{-0.05in}
 \begin{align}\label{eq::KF-Kalman-gain}
&\vect{K}_{\blue{\mathrm{EMV}}}^l=\,\begin{bmatrix}\vect{I}_{n^l}&\vect{0}\end{bmatrix}\begin{bmatrix}
 \vect{P}^{l\prpg}&\vect{P}_{lk}^{\prpg}\\
 {\vect{P}_{lk}^{\prpg}}^\top&\vect{P}^{k\prpg}
 \end{bmatrix}\bm\tilde{\vect{H}}^{i,l\top}\times\nonumber \\&~\quad \left(\bm\tilde{\vect{H}}^{i,l}\!\begin{bmatrix}
 \vect{P}^{l\prpg}&\vect{P}_{lk}^{\prpg}\\
 {\vect{P}_{lk}^{\prpg}}^\top&\vect{P}^{k\prpg}
 \end{bmatrix}\bm\tilde{\vect{H}}^{i,l\top}\!\!\!+\!\vect{R}^i\right)^{-1}\!\!\!,~k\in\!\{i,j\}\backslash\{l\}.
\end{align}

\blue{The EMV updated covaraince is $\vect{P}_{\blue{\mathrm{EMV}}}^{l\updt}=\mathrm{E}_f[(\vect{x}^l-\Hvect{x}^{l\updt})(\vect{x}^l-\Hvect{x}^{l\updt})]$ where $\vect{K}^l=\vect{K}_{\text{EMV}}^l$  $l\!\in\!\{i,j\}$ is used in~\eqref{eq::Ef_P_l}.}

\blue{Once the pose estimates of any two agents are correlated,} to keep an explicit track of the correlations, the correlated agents need to communicate at each time step to \blue{propagate} and update their cross-covariance term, regardless of whether there is a relative measurement between them. Such a requirement results in a high communication cost to implement a CL scheme and bridges our desired CL in \blue{Fig.~\ref{fig::CL_diagram}. In the following section, we set to design a relative measurement update function $\mathrm{relcorrectBelief}$ that is suitable for the CL described in Fig.~\ref{fig::CL_diagram}.}


\vspace{-0.05in}
\section{Design of update rules for $\mathrm{relcorrectBelief}$}
\blue{We consider agents $i$ and $j$ with consistent correlated local beliefs $\text{bel}^{l\prpg}(t)=(\Hvect{x}^{l\prpg}(t),\vect{P}^{l\prpg}(t))$, $l\in\{i,j\}$ aiming to process the relative measurement $\vect{z}_j^i(t)$ to correct their local beliefs in the absence of explicit knowledge about their cross-covariance $\vect{P}_{ij}^{\prpg}(t)$. 
After taking the measurement, agent $i$ sends ($\vect{z}_j^i(t)$,  $\vect{R}^i,~\text{bel}^{i\prpg}(t))$ to agent $j$ and receives $\text{bel}^{j\prpg}(t)$ from agent $j$. We assume that agent $j$ knows the measurement model $\vect{h}^i(\vect{x}^i,\vect{x}^j)$ of agent $i$ and can locally calculate $\Hvect{z}_j^i=\vect{h}^i(\Hvect{x}^{i\prpg},\Hvect{x}^{j\prpg})$ and $\vect{H}^i$ in~\eqref{eq::Jacobian_H}. In what follows we present the solutions for updating the belief of agent $i$. The same approach can be used to update the belief of agent $j$; the details are omitted for brevity. For notational simplicity, we also explain our proposed methods for when there is a single relative measurement taken by agent $i$. To process multiple concurrent relative measurements, we use sequential updating (see~\mbox{\cite[page 103]{YB-PKW-XT:11}}). That is, agent $i$ first collects the local belief of the agents that it has taken relative measurement from at time $t$. Then, it processes them via our proposed methods  one after the other by using its previously updated belief as its local belief.}

Using a fact about structured positive definite matrices in~\cite[page 207 and page 350]{RAH-CRJ:91} we can always guarantee that 
\!\!\!\!\begin{align}\label{eq::PJ-BPJ}
\begin{bmatrix}
 \vect{P}^{i\prpg}(t)&\!\!\vect{P}_{ij}^{\prpg}(t)\\
 {\vect{P}_{ij}^{\prpg}}(t)^\top&\!\!\vect{P}^{j\prpg}(t)
 \end{bmatrix}\!\leq\! \begin{bmatrix}\frac{1}{\omega}\vect{P}^{i\prpg}(t)&\vect{0}\\\vect{0}&\!\!\frac{1}{1-\omega}\vect{P}^{j\prpg}(t)
\end{bmatrix}
\!,~\omega\!\in\![0,1].
\end{align}
\blue{That is, for any value of $\vect{P}_{ij}^{\prpg}(t)$, we have a discorrelated~upper bound on the joint covariance of agents $i$ and $j$. Based on this upper bound, in the following, we propose the Discorrelated Minimum Variance (DMV) relative measurement update method that does not depend on the explicit knowledge of~$\vect{P}_{ij}^{\prpg}$. 

Let the measurement update  be~\eqref{eq::our-x_c}. Observe that for any $\vect{K}^i\in\real^{n^i\times n_z^i}$, due to~\eqref{eq::PJ-BPJ}, $\mathrm{E}_f[(\vect{x}^i-\Hvect{x}^{i\updt})(\vect{x}^i-\Hvect{x}^{i\updt})^\top]$ in~\eqref{eq::Ef_P_l} satisfies 
\begin{align}\label{eq::Ef_P_l-bound}
&\mathrm{E}_f[(\vect{x}^i-\Hvect{x}^{i\updt})(\vect{x}^i-\Hvect{x}^{i\updt})^\top]\leq \Bvect{\mathsf{P}}^{i}(\omega,\vect{K}^{i})=\nonumber \\
&\qquad\begin{bmatrix}
 (\vect{I}\!-\!\vect{K}^i\vect{H}^i_i)&-\vect{K}^i\vect{H}^i_j
 \end{bmatrix}
 \begin{bmatrix}\frac{1}{\omega}\vect{P}^{i\prpg}&\vect{0}\\\vect{0}&\frac{1}{1-\omega}\vect{P}^{j\prpg}
\end{bmatrix}\times\nonumber\\
 &\qquad\begin{bmatrix}
 (\vect{I}\!-\!\vect{K}^i\vect{H}^i_i)&-\vect{K}^i\vect{H}^i_j
 \end{bmatrix}^\top\!+\frac{1}{\gamma}\vect{K}^i\vect{R}^i\vect{K}^{i\top},
\end{align}
for any $\omega\in[0,1]$ and $\gamma\in\{1,1-\omega\}$. $\gamma=1-\omega$ is considered because it facilitates the optimization over $\omega$. Next, we let}
\begin{align}\label{eq::K_omega}
    \blue{\Bvect{\mathsf{K}}}^i(\omega)=\underset{\blue{\vect{K}^i}}{\argmin} \text{Tr}(\blue{\Bvect{\mathsf{P}}^{i}}(\omega,\vect{K}^{i})),
\end{align}
i.e., we find a gain that minimizes the total uncertainty of the upper bound $\Bvect{\mathsf{P}}^{i}(\omega,\vect{K}^{i})$ for $\omega\in[0,1]$. 
\blue{Using standard manipulations (see~\cite{JLC-JLJ:12}),} the solution of~\eqref{eq::K_omega}~is
\begin{align}\label{eq::Kl_omega}
    \blue{\Bvect{\mathsf{K}}}^i(\omega)\!=\! \frac{\vect{P}^{i\prpg}}{\omega}\vect{H}_i^{i\top}\big(\vect{H}_{i}^i\frac{\vect{P}^{i\prpg}}{\omega}\vect{H}_{i}^{i\top}\!\!\!+\!\vect{H}_{j}^i\frac{\vect{P}^{j\prpg}}{1\!-\!\omega}\vect{H}_{j}^{i^\top}\!\!\!+\!\frac{\vect{R}^i}{\gamma}\big)^{-1}\!\!\!.
\end{align}

Using this gain, $\blue{\Bvect{\mathsf{P}}}^{i}(\omega,\blue{\Bvect{\mathsf{K}}}^{i}(\omega))$ in~\eqref{eq::Ef_P_l-bound} reads as 
\begin{align}\label{eq::P_l_omega}
    &\blue{\Bvect{\mathsf{P}}}^{i}(\omega)=\blue{\Bvect{\mathsf{P}}}^{i}(\omega,\blue{\Bvect{\mathsf{K}}}^{i}(\omega))=\frac{\vect{P}^{i\prpg}}{\omega}-\frac{\vect{P}^{i\prpg}}{\omega}\vect{H}_{i}^{i\top}\times \nonumber\\
    &\qquad\qquad \big(\vect{H}_{i}^i\frac{\vect{P}^{i\prpg}}{\omega}\vect{H}_{i}^{i\top}\!\!\!\!+\!\vect{H}_{j}^i\frac{\vect{P}^{j\prpg}}{1\!-\!\omega}\vect{H}_{j}^{i\top}\!\!+\!\frac{\vect{R}^i}{\gamma}\big)^{-1}\vect{H}_{i}^i\frac{\vect{P}^{i\prpg}}{\omega},
\end{align}
which using the Matrix Inversion Lemma (\cite[page 19]{RAH-CRJ:85}) can also be expressed as,
\begin{align}\label{eq::P_l_omega_inverse}
    \blue{\Bvect{\mathsf{P}}}^{i}(\omega)=&
  \big(\omega(\vect{P}^{i\prpg})^{-1}+(1-\omega)\vect{H}_i^{i\top}(\vect{H}_j^{i}\vect{P}^{j\prpg}\vect{H}_j^{i\top}\nonumber\\&~~~~\qquad\qquad\qquad+\frac{(1-\omega)}{\gamma}\vect{R}^{i})^{-1}\vect{H}_i^{i}\big)^{-1}\!.
\end{align}
We obtain the optimal $\omega\in[0,1]$ from  
\begin{align}\label{eq::omega_star}
\omega^i_\star=\underset{0\leq\omega\leq 1}{\argmin} ~\log\det\,&\Bvect{\mathsf{P}}^{i}(\omega),
\end{align}
i.e., we obtain $\omega$ that minimizes the total uncertainty $\det(\Bvect{\mathsf{P}}^{i}(\omega))$. For $\gamma=1$, the optimization problem~\eqref{eq::P_l_omega_inverse} is not convex, but it can still be solved in an efficient manner using line search algorithms. \blue{When $\gamma=1-\omega$, $\Bvect{\mathsf{P}}^{i}(\omega)=\big(\omega(\vect{P}^{i\prpg})^{-1}\!+(1-\omega)\vect{H}_i^{i\top}(\vect{H}_j^{i}\vect{P}^{j\prpg}\vect{H}_j^{i\top}\!\!+\!\vect{R}^{i})^{-1}\vect{H}_i^{i}\big)^{-1}$ is a symmetric positive definite matrix, which depends affinely on $\omega\in[0,1]$. Therefore, the optimization problem~\eqref{eq::P_l_omega_inverse} is convex because the logarithm of the determinant of the inverse of a positive definite matrix is a convex function~\cite{LV-SB-SW:98}}. 
For $\gamma=1-\omega$, however, the updated covariance matrix will be more conservative.
\blue{We note here that we can also obtain the optimal $\omega$ from minimizing $\text{Tr}(\Bvect{\mathsf{P}}^{i}(\omega))$. In this case also, for $\gamma=1-\omega$ the optimization problem is convex.}   

Given the developments above, the \textbf{DMV updated belief} $\text{bel}^{i\updt}_{\text{DMV}}(t)=(\Hvect{x}^{i\updt}_{\text{DMV}}(t),\vect{P}^{i\updt}_{\text{DMV}}(t))$ \blue{for agent $i$} is 
\begin{subequations}\label{eq::DMV_update}
\begin{align} \Hvect{x}_{\text{DMV}}^{i\updt}&=\Hvect{x}^{i\prpg}+\blue{\vect{K}^i_{\text{DMV}}}\,(\vect{z}^i_{j}-\Hvect{z}^i_{j}),\\
\vect{P}_{\text{DMV}}^{i\updt}&=\blue{\Bvect{\mathsf{P}}}^{i}(\omega_\star^i),
\end{align}
\end{subequations}
where \blue{$\vect{K}^i_{\text{DMV}}=\Bvect{\mathsf{K}}^i(\omega_\star^i)$,} and $\Bvect{\mathsf{K}}^i(\omega_\star^i)$ and $\Bvect{\mathsf{P}}^{i}(\omega_\star)$ are given by, respectively,~\eqref{eq::Kl_omega} and~\eqref{eq::P_l_omega} evaluated at $\omega_\star^i$ of~\eqref{eq::omega_star}.
\begin{thm}\label{thm::DMV}
\blue{Given $\text{bel}^{i\prpg}(t)$, $\text{bel}^{j\prpg}(t)$ and $\vect{z}^i_j(t)$}, the DMV updated belief~\eqref{eq::DMV_update} \blue{at time $t$, for any $\gamma\in\{1,1-\omega_\star^i\}$} satisfies \begin{subequations}\label{eq::DMV}
\begin{align}
&\vect{P}^{i\updt}_{\text{DMV}}(t)\geq \text{E}_f[(\vect{x}^i(t)\!-\!\Hvect{x}^{i\updt}_{\text{DMV}}(t))(\vect{x}^i(t)\!-\!\Hvect{x}^{i\updt}_{\text{DMV}}(t))^\top]\label{eq::DMV-consistent}\\
&\det(\vect{P}^{i\updt}_{\text{DMV}}(t))\leq \det(\vect{P}^{i\prpg}(t)),\label{eq::DMV-local_improv}\\
&\vect{P}^{i\updt}_{\text{DMV}}(t)\geq\vect{P}^{i\updt}_{\text{EMV}}(t).\label{eq::DMV-EMV}
\end{align}
\end{subequations}
\end{thm}

Validity of~\eqref{eq::DMV-consistent} follows directly from~\eqref{eq::Ef_P_l-bound} (recall that~\eqref{eq::Ef_P_l-bound} holds for any $\vect{K}^i$ and \blue{any} $\omega\in[0,1] (\gamma\in\{1,1-\omega\}$). \blue{ Optimization problem~\eqref{eq::omega_star} guarantees that  $\det(\vect{P}^{i\updt}_{\text{DMV}}(t))=\det(\Bvect{\mathsf{P}}^{i}(\omega_\star^i))\leq \det(\Bvect{\mathsf{P}}^{i}(\omega))$ for any $\omega\in[0,1]$. Then,~\eqref{eq::DMV-local_improv} follows from the fact that according to~\eqref{eq::P_l_omega_inverse} we have  $\Bvect{\mathsf{P}}^{i}(\omega=1)=\vect{P}^{i\prpg}(t)$.}

\blue{Next, we validate~\eqref{eq::DMV-EMV}}. Let 
\begin{subequations}
\begin{align}
   & \vect{P}_{J,\text{EMV}}^{\updt}=\left(\begin{bmatrix}
 \vect{P}^{i\prpg}&\vect{P}_{ij}^{\prpg}\\
 {\vect{P}_{ij}^{\prpg}}^\top&\vect{P}^{j\prpg}
 \end{bmatrix}^{-1}\!\!\!\!+ \vect{H}^{i\top}\vect{R}^{i^{-1}}\vect{H}^i\right)^{-1}\!\!\!\!,\label{eq::EMV-alternative}\\
  &\bm\tilde{\vect{P}}(i,\omega)\!=\!
\left(\!\begin{bmatrix}\frac{1}{\omega}\vect{P}^{i\prpg}&\vect{0}\\\vect{0}&\!\frac{1}{1-\omega}\vect{P}^{j\prpg}
\end{bmatrix}^{-1}\!\!\!\!\!\!+\vect{H}^{i\top}\big(\frac{\vect{R}^i}{\gamma}\big)^{-1}\vect{H}^i\!\right)^{-1}\!\!\!\!\!.\label{eq::P_tilde}
\end{align} 
\end{subequations}

Using standard manipulations, we  \blue{can} show that EMV and DMV updated covariance matrices are
\begin{subequations}
\begin{align}
    \vect{P}_{\text{EMV}}^{i\updt}&=\begin{bmatrix}\vect{I}_{n^i}&\vect{0}\end{bmatrix} \vect{P}_{J,\text{EMV}}^{\updt}\begin{bmatrix}\vect{I}_{n^i}\\\vect{0}\end{bmatrix},\label{eq::proof_EMV_updt}\\
  \vect{P}_{\text{DMV}}^{i\updt}&=\blue{\Bvect{\mathsf{P}}}^{i}(\omega_\star^i)=\begin{bmatrix}\vect{I}_{n^i}&\vect{0}\end{bmatrix} \bm\tilde{\vect{P}}(i,\omega^\star)\begin{bmatrix}\vect{I}_{n^i}\\\vect{0}\end{bmatrix}.\label{eq::proof_DMV_Pw}
\end{align}
\end{subequations}

Note that $   \vect{P}_{J,\text{EMV}}^{\updt^{-1}}\!-\!\bm\tilde{\vect{P}}(i,\omega)^{-1}=(1\!-\!\gamma)\,\vect{H}^{i\top}\vect{R}^{i^{-1}}\vect{H}^i\!+\!
   \left[\begin{smallmatrix}
 \vect{P}^{i\prpg}&\vect{P}_{ij}^{\prpg}\\
 {\vect{P}_{ij}^{\prpg}}^\top&\vect{P}^{j\prpg}
 \end{smallmatrix}\right]^{-1}\!\!\!\!-\left[\begin{smallmatrix}\frac{1}{\omega}\vect{P}^{i\prpg}&\vect{0}\\\vect{0}&\frac{1}{1-\omega}\vect{P}^{j\prpg}
\end{smallmatrix}\right]^{-1}\!\!,$
which by virtue of~\eqref{eq::PJ-BPJ}, guarantees that $\vect{P}_{J,\text{EMV}}^{\updt}(t)^{-1}-\bm\tilde{\vect{P}}(i,\omega)^{-1}\geq0$ or equivalently $\vect{P}_{J,\text{EMV}}^{\updt}(t)\leq \bm\tilde{\vect{P}}(i,\omega)$ for all $\omega\in[0,1]$ and $\gamma\in\{1,1-\omega\}$.  Then, $\vect{P}_{\text{DMV}}^{i\updt}(t)\geq\vect{P}_{\text{EMV}}^{i\updt}(t)$ follows from~\eqref{eq::proof_EMV_updt} and~\eqref{eq::proof_DMV_Pw}.

\rm{\eqref{eq::DMV-consistent} indicates that, despite the lack of knowledge about~$\vect{P}_{ij}^{\prpg}$, the  DMV update is consistent in first-order approximate sense. \eqref{eq::DMV-local_improv} shows that the DMV update is guaranteed to be no worse than the agent's local belief, while~\eqref{eq::DMV-EMV} indicates that DMV, as one expects, does not out-perform EMV. 
}

\begin{rem}[Joint DMV update]\rm{
Let $\vect{K}_{\text{EMV}}^l$ be the EMV gain of agents  $l\in\{i,j\}$ computed from~\eqref{eq::K-opt-indiv}. Evidently, these gains satisfy 
$[\vect{K}_{\text{EMV}}^{i\top}~\vect{K}_{\text{EMV}}^{j\top}]^{\top}=\underset{\blue{\vect{K}}}{\argmin}\text{Tr}(\vect{P}^{\updt}_{J})$, where $\vect{P}^{\updt}_{J}=E_f[(\vect{x}_J-\Hvect{x}^{\updt}_J)(\vect{x}_J-\Hvect{x}^{\updt}_J)^\top]=(\vect{I}-\vect{K}\vect{H}^i)\left[\begin{smallmatrix}\vect{P}^{i\prpg}&\vect{P}_{ij}^{\prpg}\\
{\vect{P}_{ij}^{\prpg}}^\top&\vect{P}^{j\prpg}\end{smallmatrix}\right](\vect{I}-\vect{K}\vect{H}^i)^\top\!\!\!+\!\vect{K}\vect{R}^i\vect{K}^\top$ i.e., they minimize the joint updated covartiance matrix. Under this observation, in our preliminary work in~\cite{JZ-SSK:17}, we pursued an alternative DMV design that computed the update gain for the agents $\{i,j\}$ jointly, i.e., we obtained $[\vect{K}_{\text{DMV}}^{i\top}~\vect{K}_{\text{DMV}}^{j\top}]^{\top}=\argmin\text{det}(\Bvect{P}^{\updt}_J)$, where $\Bvect{P}^{\updt}_J=(\vect{I}\!-\!\vect{K}\vect{H}^i)\left[\begin{smallmatrix}\frac{1}{\omega}\vect{P}^{i\prpg}&\vect{0}\\\vect{0}&\frac{1}{1-\omega}\vect{P}^{j\prpg}
\end{smallmatrix}\right](\vect{I}\!-\!\vect{K}\vect{H}^i)^\top\!\!\!+\!\vect{K}\vect{R}^i\vect{K}^\top$,  $\omega\in[0,1]$. For any $\omega\in[0,1]$, we have $\Bvect{P}^{\updt}_J\geq \vect{P}^{\updt}_{J}$. ~\cite{JZ-SSK:17} shows that the optimal $\omega$ can be obtained from a convex optimization problem. This alternative DMV method results in updated estimates that satisfy~\eqref{eq::DMV-consistent} and ~\eqref{eq::DMV-EMV}. However,~\eqref{eq::DMV-local_improv} does not necessarily hold. That is, one cannot guarantee that the updated estimates will be better than the local ones. In fact, numerical examples shows that~\eqref{eq::DMV-local_improv} can be violated.
}
\end{rem}

\subsection{An alternative solution for $\mathrm{relcorrectBelief}$}
Since DMV's upper bound on the joint covariance matrix accounts for all the possible values for the unknown cross-covariance matrix, the DMV updates generally are  too conservative.  The Estimated Cross-covariance Minimum Variance (ECMV) update method, which we devise next as an alternative function for $\mathrm{relcorrectBelief}$, aims to reduce this conservatism by estimating the unknown cross-covariance matrix. 

Let the joint covariance matrix of agents $\{i,j\}$ at time $t$, prior to processing $\vect{z}_j^i(t)$ be
\begin{align}\label{eq::P_JX}
\vect{P}_{J}^{\prpg}(\vect{X})=\begin{bmatrix}
 \vect{P}^{i\prpg}(t)&\vect{X}(t)\\
\vect{X}(t)^\top&\vect{P}^{j\prpg}(t)
 \end{bmatrix}\geq 0,
 \end{align}
  where~$\vect{X}$ indicates the unknown cross-covariance matrix.
  In lieu of~\eqref{eq::Ef_P_l} and~\eqref{eq::K-opt-indiv} in the EMV design, ECMV design estimates the unknown cross-covariance matrix $\vect{X}$~in
\begin{align}\label{eq::our-P-kform-indv-X}
\vect{P}^{i\updt}(\vect{K}^i,\vect{X})=&\begin{bmatrix}(\vect{I}\!-\!\vect{K}^i\vect{H}_i^i)&-\vect{K}^i\vect{H}_j^i
\end{bmatrix}\begin{bmatrix}
 \vect{P}^{i\prpg}&\vect{X}\\
 \vect{X}^\top&\vect{P}^{j\prpg}
 \end{bmatrix}
\times\nonumber\\
&\begin{bmatrix}(\vect{I}\!-\!\vect{K}^i\vect{H}_i^i)&-\vect{K}^i\vect{H}_j^i
\end{bmatrix}^\top
\!\!\!+\!\vect{K}^i\vect{R}^i\vect{K}^{i\top}\!\!,
\end{align}
from the following optimization problem
\begin{subequations}\label{eq::ECMV-opt}
\begin{align}
    &(\vect{K}^{i\star},\vect{X}^\star)=\arg\underset{\vect{K}^i}{\min}\,\underset{\vect{X}}{\max}\, \text{Tr}\big(\vect{P}^{i\updt}(\vect{K}^i,\vect{X})\big),\label{eq::ECMV-opt-Objective}\\
   &\text{subject to}~ \begin{bmatrix}
 \vect{P}^{i\prpg}&\vect{X}\\
\vect{X}^\top&\vect{P}^{j\prpg}
 \end{bmatrix}\geq 0.\label{eq::ECMV-opt-const}
\end{align}
\end{subequations}
The idea here is to (conservatively) estimate the unknown $\vect{P}_{ij}^{\prpg}$ by finding an $\vect{X}^\star$ that gives the most conservative updated covariance, and then find a gain $\vect{K}^{i\star}$ that minimizes this conservative updated covariance. This approach is along the line of~\cite{SL-KD:17}. Here, we provide a rigorous proof for the consistency of this design in trace sense and also we show how this design compares to the EMV and DMV updates. Moreover, since the optimization problem~\eqref{eq::ECMV-opt} is numerically expensive, we also present a \emph{practical} alternative method, which trades in guaranteed trace consistency of the ECMV design in first order sense for a tractable numerical solution procedure that maintains the other properties of the ECMV~method.

Given the developments above, the \textbf{ECMV updated belief}  $\text{bel}^{i\updt}_{\text{ECMV}}(t)=(\Hvect{x}^{i\updt}_{\text{ECMV}}(t),\vect{P}^{i\updt}_{\text{ECMV}}(t))$ for agent $i$ is 
\begin{subequations}\label{eq::ECMV_update}
\begin{align} \Hvect{x}_{\text{ECMV}}^{i\updt}&=\Hvect{x}^{i\prpg}+\blue{\vect{K}^{i}_{\text{ECMV}}}\,(\vect{z}^i_{j}-\Hvect{z}^i_{j}),\\
\vect{P}_{\text{ECMV}}^{i\updt}&=\vect{P}^{i\updt}(\vect{K}^{i\star},\vect{X}^\star),
\end{align}
\end{subequations}
where \blue{$\vect{K}^{i}_{\text{ECMV}}=\vect{K}^{i\star}$} and $(\vect{K}^{i\star},\vect{X}^\star)$ is an optimal solution of~\eqref{eq::ECMV-opt}. 

\begin{thm}\label{thm::ECMV1}
\rm{\blue{Given $\text{bel}^{i\prpg}(t)$, $\text{bel}^{j\prpg}(t)$ and $\vect{z}^i_j(t)$}, the ECMV updated belief~\eqref{eq::ECMV_update} \blue{at time $t$} satisfies
\begin{subequations}
\label{eq::ECMV-property}
\begin{align}
&\text{Tr}(\vect{P}^{i\updt}_{\text{ECMV}}(t))\geq\nonumber\\ 
&~~\quad\text{Tr}\big(E_f[(\vect{x}^i(t)-\Hvect{x}^{i\updt}_{\text{ECMV}}(t))(\vect{x}^i(t)-\Hvect{x}^{i\updt}_{\text{ECMV}}(t))^\top]\big),\label{eq::ECMV-consistent-trace}\\
&\text{Tr}(\vect{P}_{\text{ECMV}}^{i\updt}(t))\leq \text{Tr}(\vect{P}^{i\prpg}(t)),\label{eq::ECMV-local-improv-trace}\\
&\text{Tr}(\vect{P}_{\text{EMV}}^{i\updt}(t))\leq \text{Tr}(\vect{P}_{\text{ECMV}}^{i\updt}(t))\leq \text{Tr}(\vect{P}_{\text{DMV}}^{i\updt}(t)).\label{eq::ECMV-over-EMV}
\end{align}
\end{subequations}
Moreover, if $\Big[\begin{smallmatrix}
 \vect{P}^{i\prpg}&\vect{X}^\star\\
\vect{X}^{\star\top}&\vect{P}^{j\prpg}
 \end{smallmatrix}\Big]>0$, we have
 \begin{subequations}
\begin{align}
    &\vect{P}^{i\updt}_{\text{ECMV}}(t)\leq \vect{P}^{i\prpg}(t),\label{eq::ECMV-local-improv}\\
&\vect{P}_{\text{ECMV}}^{i\updt}(t)\leq \vect{P}_{\text{DMV}}^{i\updt}(t).\label{eq::ECMV-better-DMV}
\end{align}
\end{subequations}
}
\end{thm}

\blue{Since the objective function of~\eqref{eq::ECMV-opt} for every fix $\vect{K}^i$ is concave in $\vect{X}$ and for every fixed $\vect{X}$ is convex in $\vect{K}^i$,~\eqref{eq::ECMV-opt} is a convex-concave optimization problem.} 
Let $\mathcal{X}=\big\{\vect{X}\in\real^{n^i\times n^j}\,\big|\,\left[\begin{smallmatrix}\vect{P}^i&\vect{X}\\\vect{X}^\top&\vect{P}^j\end{smallmatrix}\right]\geq 0\big\}$. The set $\mathcal{X}$ is convex and compact. Therefore, using Sion's minmax result~\cite[Corollary 3.3]{MS:58}, we have the guarantees that 
\begin{align*}
\underset{\vect{K}^i\in\real^{n^i\times n_z^i}}{\min}\,\,\underset{\vect{X}\in\mathcal{X}}{\max} \text{Tr}\big(\vect{P}^{i\updt}(\vect{K}^i,&\vect{X})\big)=\\
&\underset{\vect{X}\in\mathcal{X}}{\max}\,\,\underset{\vect{K}^i\in\real^{n^i\times n_z^i}}{\min} \text{Tr}\big(\vect{P}^{i\updt}(\vect{K}^i,\vect{X})\big),
\end{align*} 
and the optimization problem~\eqref{eq::ECMV-opt} satisfies
\begin{align}\label{eq::saddle}
    \text{Tr}\big(\vect{P}^{i\updt}(\vect{K}^{i\star},&\vect{X})\big)\leq\text{Tr}\big(\vect{P}^{i\updt}(\vect{K}^{i\star},\vect{X}^{\star})\big)\leq \text{Tr}\big(\vect{P}^{i\updt}(\vect{K}^i,\vect{X}^\star)\big).
\end{align}
 Then, the validity of~\eqref{eq::ECMV-consistent-trace} and~\eqref{eq::ECMV-local-improv-trace} follows from the facts that, respectively, $E_f[(\vect{x}^i(t)-\Hvect{x}^{i\updt}_{\text{ECMV}}(t))(\vect{x}^i(t)-\Hvect{x}^{i\updt}_{\text{ECMV}}(t))^\top]=\vect{P}^{i\updt}(\vect{K}^{i\star},\vect{X}=\vect{P}_{ij}^{\prpg})$ and $\vect{P}^{i\prpg}=\vect{P}^{i\updt}(\vect{K}^{i}=\vect{0},\vect{X}^{\star})$. 
We validate~\eqref{eq::ECMV-over-EMV} as follows. Note that \begin{align}\label{eq::intermidate}
    \underset{\vect{X}\in\mathcal{X}}{\max}\,\,\underset{\vect{K}^i\in\real^{n^i\times n_z^i}}{\min} \text{Tr}\big(\vect{P}^{i\updt}(\vect{K}^i,\vect{X})\big)=\underset{\vect{X}\in\mathcal{X}}{\max}\,\text{Tr}\big(\vect{P}^{i\updt}(\vect{K}^i(\vect{X}),\vect{X})\big),
\end{align}
where $\vect{K}^{i}(\vect{X})$ that minimizes $\text{Tr}\big(\vect{P}^{i\updt}(\vect{K}^i,\vect{X})\big)$ is 
\begin{align}\label{eq::K_ECMV-gain}
\vect{K}^{i}(\vect{X})=&\begin{bmatrix}\vect{I}_{n^i}&\vect{0}\end{bmatrix}\,\vect{P}_J^{\prpg}(\vect{X})\vect{H}^{i\top}_{i}\!(\vect{H}^{i}_i\vect{P}_J^{\prpg}(\vect{X})\vect{H}^{i\top}_i\!\!\!+\!\vect{R}^i)^{-1}.
\end{align}
Therefore, $\vect{P}^{i\updt}(\vect{K}^i(\vect{P}_{ij}^{\prpg}),\vect{P}_{ij}^{\prpg})=\vect{P}_{\text{EMV}}^{i\updt}$. Since $\vect{X}=\vect{P}_{ij}^{\prpg}$ is in the feasible set of $\underset{\vect{X}\in\mathcal{X}}{\max}\,\text{Tr}\big(\vect{P}^{i\updt}(\vect{K}^i(\vect{X}),\vect{X})\big)$, we have the guarantees that $\text{Tr}(\vect{P}_{\text{EMV}}^{i\updt}(t))\leq \text{Tr}(\vect{P}_{\text{ECMV}}^{i\updt}(t))$, validating the lower bound in~\eqref{eq::ECMV-over-EMV}.  Let $\vect{K}_{\text{DMV}}^i$ be the update gain of the DMV update method for agent $i$. Given~\eqref{eq::Ef_P_l} and~\eqref{eq::our-P-kform-indv-X} along with~\eqref{eq::PJ-BPJ} we can write, for any $\omega\in[0,1]$ and $\gamma\in\{1,1-\omega\}$,
\begin{align*}
    &\vect{P}^{i\updt}(\vect{K}_{\text{DMV}}^i,\vect{X}^\star)-\vect{P}^{i\updt}_{\text{DMV}}=\big(1-\frac{1}{\gamma}\big)\vect{K}_{\text{DMV}}^i\vect{R}^i\vect{K}_{\text{DMV}}^{i\top}+\\
    &
    \begin{bmatrix}
 (\vect{I}\!-\!\vect{K}_{\text{DMV}}^i\vect{H}^i_i)&-\vect{K}_{\text{DMV}}^i\vect{H}^i_j
 \end{bmatrix}
 \left(\begin{bmatrix}
 \vect{P}^{i\prpg}&\vect{X}^\star\\
 {\vect{X}^\star}^\top&\vect{P}^{j\prpg}
 \end{bmatrix}-\right.
 \\
 &\left.\begin{bmatrix}\frac{1}{\omega}\vect{P}^{i\prpg}&\vect{0}\\\vect{0}&\frac{1}{1-\omega}\vect{P}^{j\prpg}
\end{bmatrix}\right)
 \begin{bmatrix}
 (\vect{I}\!-\!\vect{K}_{\text{DMV}}^i\vect{H}^i_i)&-\vect{K}_{\text{DMV}}^i\vect{H}^i_j
 \end{bmatrix}^\top\!\!\leq 0.
\end{align*}
Because  $\vect{P}^{i\updt}(\vect{K}_{\text{DMV}}^i,\vect{X}^\star)\leq \vect{P}^{i\updt}_{\text{DMV}}$, by virtue of~\eqref{eq::saddle} we can guarantee that the upper bound in~\eqref{eq::ECMV-over-EMV} holds. Next, let $(\vect{K}^{i\star},\vect{X}^\star)$ be a solution of~\eqref{eq::ECMV-opt} which satisfies
$\vect{P}_J^{\prpg}(\vect{X}^\star)=\Big[\begin{smallmatrix}
 \vect{P}^{i\prpg}&\vect{X}^\star\\
\vect{X}^{\star\top}&\vect{P}^{j\prpg}
 \end{smallmatrix}\Big]>0$.
Then, by substituting gain~\eqref{eq::K_ECMV-gain}  in~\eqref{eq::our-P-kform-indv-X}, and after some standard matrix inversion manipulations we obtain
\begin{align}\label{eq::P-bar}
    \vect{P}^{i\updt}(\vect{K}^{i\star},\vect{X}^\star)=\begin{bmatrix}\vect{I}_{n^i}&\vect{0}
    \end{bmatrix}\bar{\vect{P}} \begin{bmatrix}\vect{I}_{n^i}\\\vect{0}
    \end{bmatrix}.
\end{align}
where $\Bvect{P}=\Big(\vect{P}_J^{\prpg}(\vect{X}^{\star})^{-1}\!\!+\vect{H}^{i\top}_{i}\!\vect{R}^{i^{-1}}\vect{H}^{i}_i\Big)^{-1}$. Since $\bar{\vect{P}}^{-1}\geq \vect{P}_J^{\prpg}(\vect{X}^\star)^{-1}$, we have $\bar{\vect{P}}\leq \vect{P}_J^{\prpg}(\vect{X}^\star)$. Subsequently, we conclude that~\eqref{eq::ECMV-local-improv} holds. Next,we validate~\eqref{eq::ECMV-better-DMV}.  Recall $\bm\tilde{\vect{P}}(i,\omega)$ defined in~\eqref{eq::P_tilde}. For any $\omega\in[0,1]$ and $\gamma\in\{1,1-\omega\}$, we can write $\bm\bar{\vect{P}}^{-1}-\bm\tilde{\vect{P}}(i,\omega)^{-1}=(1-\gamma)\,\vect{H}^{i\top}\vect{R}^{i^{-1}}\vect{H}^i+\left[
   \begin{smallmatrix}
 \vect{P}^{i\prpg}&\vect{X}^\star\\
 {\vect{X}^\star}^\top&\vect{P}^{j\prpg}
 \end{smallmatrix}\right]^{-1}\!\!\!\!-\left[\begin{smallmatrix}\frac{1}{\omega}\vect{P}^{i\prpg}&\vect{0}\\\vect{0}&\frac{1}{1-\omega}\vect{P}^{j\prpg}
\end{smallmatrix}\right]^{-1}\!\!\!\!,$
which by virtue of~\eqref{eq::PJ-BPJ}, guarantees that $\bm\bar{\vect{P}}^{-1}-\bm\tilde{\vect{P}}(i,\omega)^{-1}\geq0$ or equivalently $\bar{\vect{P}}\leq \bm\tilde{\vect{P}}(i,\omega)$. Then, $\vect{P}_{\text{ECMV}}^{i\updt}(t)\leq \vect{P}_{\text{DMV}}^{i\updt}(t)$ follows from~\eqref{eq::proof_DMV_Pw} and~\eqref{eq::P-bar}.
\blue{The properties~\eqref{eq::ECMV-over-EMV} and~\eqref{eq::ECMV-better-DMV} indicate that the ECMV method, by estimating the unknown cross-covariance, delivers a better result than the DMV update.}

\emph{A practical ECMV update procedure}: since the optimization problem~\eqref{eq::ECMV-opt} of the ECMV is numerically expensive, in the following we present an alternative method with a less numerical cost. We refer to this alternative design as \emph{practical} ECMV, or PECMV for short. The idea here is to use the update gain~\eqref{eq::K_ECMV-gain} but instead of maximizing~$\text{Tr}\big(\vect{P}^{i\updt}(\vect{K}^i(\vect{X}),\vect{X})$ in~\eqref{eq::intermidate}, we maximize $\text{det}\big(\vect{P}^{i\updt}(\vect{K}^i(\vect{X}),\vect{X})$. In doing so, as we show below, we can estimate the unknown $\vect{X}$ from a convex linear matrix inequality optimization, for which efficient numerical solvers exists. Recall that we showed in the proof of Theorem~\ref{thm::ECMV1} that after substituting for the gain~\eqref{eq::K_ECMV-gain} and some standard manipulations (see derivation of~\eqref{eq::P-bar}) we obtain  
$ \vect{P}^{i\updt}(\vect{K}^i(\vect{X}),\vect{X})=\begin{bmatrix}\vect{I}_{n^i}&\vect{0}
    \end{bmatrix}(\vect{P}_J^{\prpg}(\vect{X})^{-1}\!\!+\vect{H}_{i}^{i\top}\!\vect{R}^{i^{-1}}\vect{H}^{i}_{i})^{-1}\begin{bmatrix}\vect{I}_{n^i}&\vect{0}
    \end{bmatrix}^\top$.
    Then, in PECMV update, we obtain $\vect{X}^\star$ from
\begin{subequations}\label{eq::PECMV-opt}
\begin{align}
    &\vect{X}^\star\!=\arg\underset{\vect{X}}{\max}\, \text{det}
    \begin{bmatrix}\vect{I}_{n^i}\\\vect{0}
    \end{bmatrix}^\top\!\!\!\!(\vect{P}_J^{\prpg}(\vect{X})^{-1}\!\!+\vect{H}^{i\top}\!\vect{R}^{i^{-1}}\vect{H}^i)^{-1}\!\begin{bmatrix}\vect{I}_{n^i}\\\vect{0}
    \end{bmatrix},\label{eq::PECMV-opt-Objective}\\
   &\text{subject to}~ \begin{bmatrix}
 \vect{P}^{i\prpg}&\vect{X}\\
\vect{X}^\top&\vect{P}^{j\prpg}
 \end{bmatrix}> 0.\label{eq::PECMV-opt-const}
\end{align}
\end{subequations}
Following~\cite[Corollary 1]{KKK:15}, the optimization problem~\eqref{eq::PECMV-opt} can be cast in the equivalent linear matrix inequality optimization
\begin{subequations}\label{eq::PECMV-opt-convex}
 \begin{align}
    &(\vect{X}^\star,\vect{Z}^\star)=\arg\underset{\vect{X},\vect{Z}}{\min}\,\text{log\,det} (\vect{Z}^{-1}),~~~\text{subject to}\\
   &\begin{bmatrix}\vect{P}^{i\prpg}-\vect{Z}&\begin{bmatrix}
    \vect{P}^{i\prpg}&\vect{X}
    \end{bmatrix}\vect{Q}^{i}\\\vect{Q}^{i}\begin{bmatrix}
    \vect{P}^{i\prpg}&\vect{X}
    \end{bmatrix}^{\top}&\vect{I}+\vect{Q}^{i}\vect{P}^{\prpg}_J(\vect{X})\vect{Q}^{i}\end{bmatrix}\geq 0,\\
    &\begin{bmatrix}
 \vect{P}^{i\prpg}&\vect{X}\\
\vect{X}^\top&\vect{P}^{j\prpg}
 \end{bmatrix}> 0,
 \end{align}
 \end{subequations}
where $\vect{Q}^{i}=\sqrt{\vect{H}^{i\top}_{i}\!\vect{R}^{i^{-1}}\vect{H}^{i}_{i}}$. To obtain~\eqref{eq::PECMV-opt-convex}, we defined the auxiliary matrix $\vect{Z}$ that is set to satisfy   $\vect{Z}\leq \begin{bmatrix}\vect{I}_{n^i}&\vect{0}
  \end{bmatrix}(\vect{P}_J^{\prpg}(\vect{X}^\star)^{-1}\!\!+\vect{H}^{i\top}_{i}\!\vect{R}^{i^{-1}}\vect{H}^{i}_{i})^{-1}\begin{bmatrix}\vect{I}_{n^i}&\vect{0}
    \end{bmatrix}^\top$.

Given the developments above, the \textbf{PECMV updated belief}   $\text{bel}^{i\updt}_{\text{PECMV}}(t)=(\Hvect{x}^{i\updt}_{\text{PECMV}}(t),\vect{P}^{i\updt}_{\text{PECMV}}(t))$ for agent $i$ is 
\begin{subequations}\label{eq::PECMV_update}
\begin{align} \Hvect{x}_{\text{PECMV}}^{i\updt}&=\Hvect{x}^{i\prpg}+\blue{\vect{K}_{\text{PECMV}}^i}\,(\vect{z}^i_{j}-\Hvect{z}^i_{j}),\\
\vect{P}_{\text{PECMV}}^{i\updt}&=\vect{P}^{i\updt}(\vect{K}^{i\star},\vect{X}^\star),
\end{align}
\end{subequations}
where \blue{$\vect{K}_{\text{PECMV}}^i=\vect{K}^{i\star}$}, $\vect{X}^\star$ is an optimal solution of~\eqref{eq::PECMV-opt-convex} and $\vect{K}^{i\star}$ is~\eqref{eq::K_ECMV-gain}~evaluated at $\vect{X}^\star$.

\begin{thm}\label{thm::PECMV}
\blue{Given $\text{bel}^{i\prpg}(t)$, $\text{bel}^{j\prpg}(t)$ and $\vect{z}^i_j(t)$}, the ECMV updated belief~\eqref{eq::PECMV_update} \blue{at time $t$} satisfies
\begin{subequations}
\label{eq::PECMV-property}
\begin{align}
    &\vect{P}^{i\updt}_{\text{PECMV}}(t)\leq \vect{P}^{i\prpg}(t),\label{eq::PECMV-local-improv}\\
&\vect{P}_{\text{PECMV}}^{i\updt}(t)\leq \vect{P}_{\text{DMV}}^{i\updt}(t),\label{eq::PECMV-better-DMV}\\
&\text{det}(\vect{P}_{\text{EMV}}^{i\updt}(t))\leq \text{det}(\vect{P}_{\text{\blue{PECMV}}}^{i\updt}(t))\leq \text{det}(\vect{P}_{\text{DMV}}^{i\updt}(t)),\label{eq::PECMV-over-EMV}
\end{align}
\end{subequations}
\end{thm}

The proof of~\eqref{eq::PECMV-local-improv} and~\eqref{eq::PECMV-better-DMV} is the same as the proof of~\eqref{eq::ECMV-local-improv} and~\eqref{eq::ECMV-better-DMV}
in Theorem~\ref{thm::ECMV1}.
 Since $\vect{P}^{i\updt}(\vect{K}^i(\vect{P}_{ij}^{\prpg}),\vect{P}_{ij}^{\prpg})=\vect{P}_{\text{EMV}}^{i\updt}$,  $\vect{X}=\vect{P}_{ij}^{\prpg}$ is in the feasible set of the optimization problem~\eqref{eq::PECMV-opt}. Consequently, $\text{det}(\vect{P}_{\text{EMV}}^{i\updt}(t))\leq \text{det}(\vect{P}_{\text{PECMV}}^{i\updt}(t))$, which validates the lower bound in~\eqref{eq::PECMV-over-EMV}. The upper bound in~\eqref{eq::PECMV-over-EMV} is deduced from~\eqref{eq::PECMV-better-DMV}. 
 
 The consistency evaluations so far were in the first-order approximate sense, \blue{used to guide our designs}. For nonlinear systems, it is customary to assess the estimation filter consistency using Monte Carlo based statistical tests such as the Normalized Estimation Error Squared (NEES)~\cite{YBS-XRL-TK:01} or the Average Normalized Estimation Error Squared (ANEES)~\cite{YBS-XRL-TK:01}; \blue{see e.g.,~\cite{TB-JN-JG-MS-EN:06,YY-GH:17}}. Next, we use a numerical example to assess and compare the consistency of our proposed DMV and PECMV update methods using the NEES~measure.

\section{Demonstrative examples}\label{sec::simulation}
We test the localization performance of a CL algorithm implementing the DMV- and PECMV-based $\mathrm{relcorrectBelief}$ in simulation and experiment for $3$ mobile robots moving on a flat terrain. \blue{The  results of the experiment, via Turtlebot robots, are  available in the video attachment of the paper~\cite{VD:19}.}

\begin{figure}[t]
\begin{center}
\includegraphics[width=3.41in,height=1.6in]{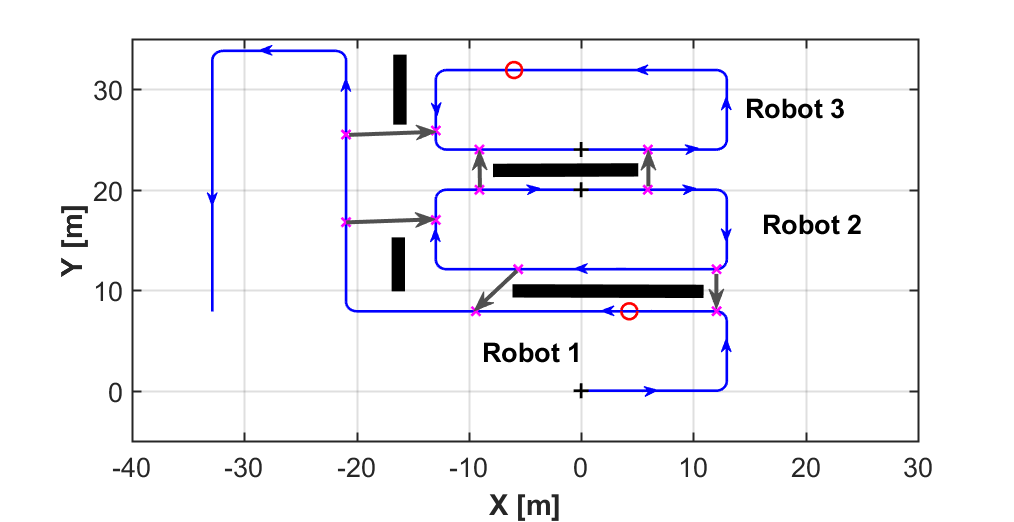}
\end{center}
\vspace{-0.1in}
\caption{\blue{\small The true trajectories of the robots in the simulation: the robots start at $+$ and return to it following their trajectories. The $\times$'s show the location of the relative measurements with the arrow showing which robot has taken the relative measurement (at each incidence only one relative measurement is taken). The red $\textup{o}$ shows the location that a robot obtains an absolute~measurement.
}}\vspace{-0.2in}
\label{Fig::traj-simul}
\end{figure}

\emph{Simulation}: 
The equations of motion of the robots, given their linear velocity $v^i(t)$ and angular velocity $\omega^i(t)$ $i\in\{1,2,3\}$, are described by,  $x^i(t+1)=x^i(t)+\Delta t\, (v_m^i(t)\cos(\phi^i(t)))$,
    $y^i(t+1)=x^i(t)+\Delta t\, (v_m^i(t)\sin(\phi^i(t)))$,
    $\phi^i(t+1)=\phi^i(t)+\Delta t \,\omega_m^i(t)$,
where $v_m^i(t)=v^i(t)+\nu_v^i(t)$ and $\omega^i_m(t)=\omega^i(t)+\nu_{\omega}^i(t)$.
Here $v^i_m$ and $\omega_m^i$ are measured linear and angular velocities, while $\nu_v$ and $\nu_\omega$ are the corresponding contaminating measurement noises. The measurement noise of agents \{1,2,3\} respectively are assumed to be \{20\%,25\%,30\%\} of the linear (resp. \{20\%,15\%,10\%\} of the angular) velocity. The relative measurement used in the simulation is relative pose corrupted by relative measurement noise with standard deviation $[0.1m,0.1m,5^{\circ}]^{\top}$. Absolute range measurement, corrupted by absolute measurement noise with standard deviation $0.2m$, can be obtained occasionally with respect to landmarks that have known positions.
\blue{For the simulation scenario shown in Fig.~\ref{Fig::traj-simul},} the position root mean square error (RMSE) and the NEES are calculated from $M=50$ Monte Carlo runs. These runs provide $M$ independent samples of the position estimation error $\vect{e}^{i}_\ell(t)=[x^{i}_\ell(t),y^{i}_\ell(t)]^\top-[\hat{x}^{i}_\ell(t),\hat{y}^{i}_\ell(t)]^\top$, where $(x^{i}_\ell(t),y^{i}_\ell(t))$ is the true position and $(\hat{x}^{i}_l(t),\hat{y}^{i}_l(t))$ is the estimated position of agent $i$ with the associated error covariance matrix $\vect{P}^{i}_l(t)\in\real^{2\times 2}$ at the Monte Carlo run $\ell\until M$. 
Under the hypothesis $\mathcal{H}$ that the estimate is consistent under the Gaussian assumption,  $M\Bvect{\epsilon}^i(t)$ will be chi-square distributed with $2M$ degrees of freedom. The estimate is consistent if $\Bvect{\epsilon}^i(t)\in[r_1,r_2]$ such that $P\{\Bvect{\epsilon}^i(t)\in[r_1,r_2]|\mathcal{H}\}=1-\alpha$. The two-sided $95\%$ ($\alpha=0.05$) region for a $2M\!=\!100$~degrees of freedom chi-square distribution divided by $M$ is $[r_1,r_2]\!=\![\frac{\chi^2_{100}(0.025)}{M},\frac{\chi^2_{100}(0.975)}{M}]\!=\![1.48,2.59]$, see~\cite{YBS-XRL-TK:01} for~details. \blue{A NEES measure above $r_2$ means that the actual estimation uncertainty is much larger than what the estimator believes, while a NEES measure below $r_1$ means the opposite.
}

\begin{figure}[t]
\subfloat[Robot 1]{\includegraphics[scale=0.2]{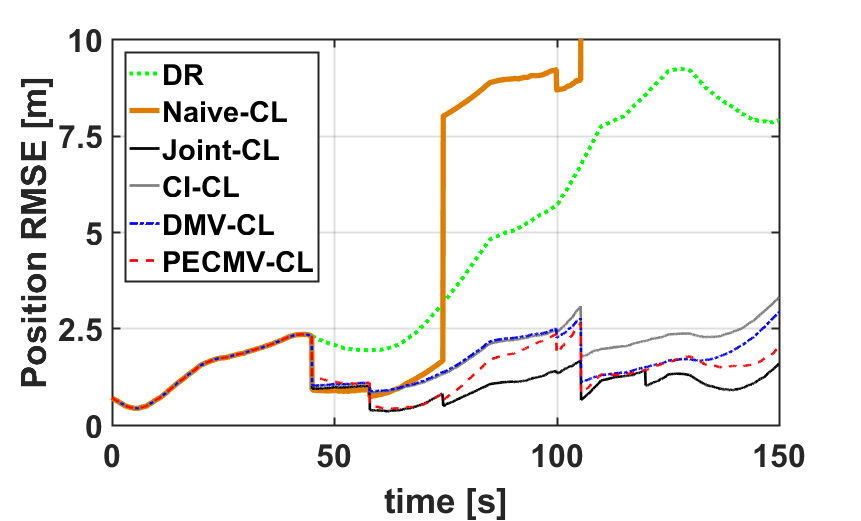}}
\subfloat[Robot 2]{\includegraphics[scale=0.2]{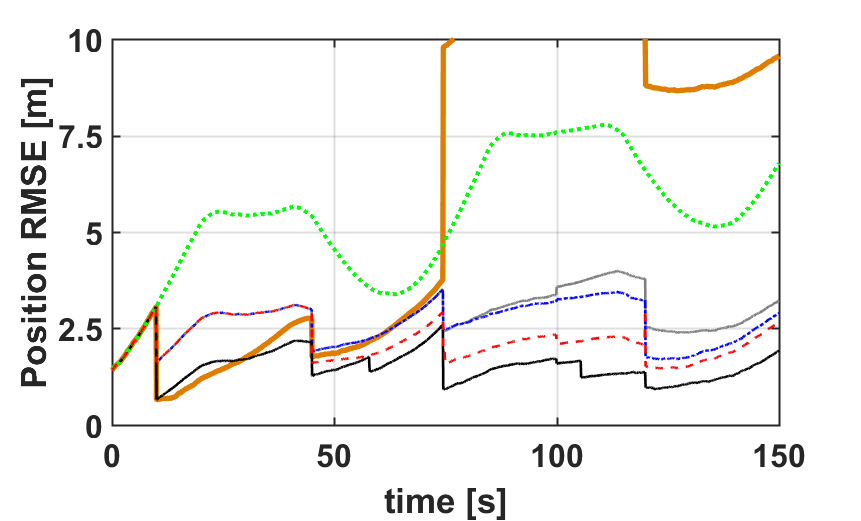}}\\\vspace{-0.1in}
\subfloat[Robot 3]{\includegraphics[scale=0.2]{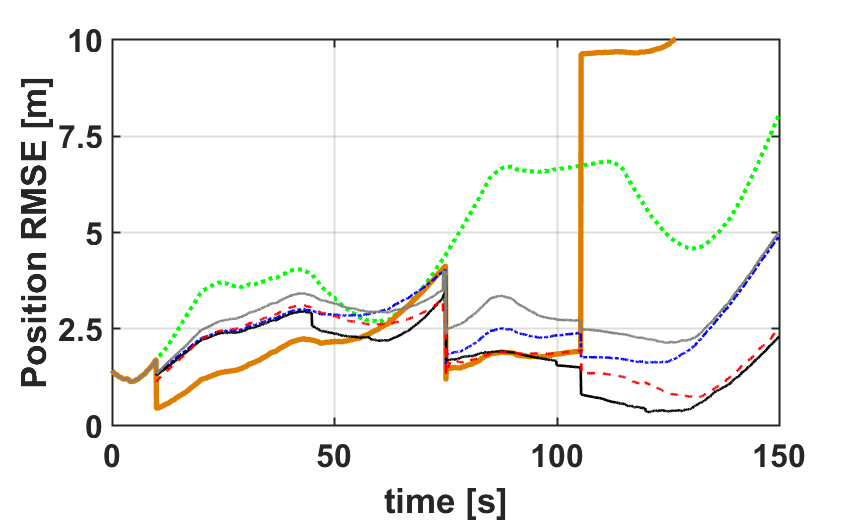}}
\subfloat[Robot 1]{\includegraphics[scale=0.2]{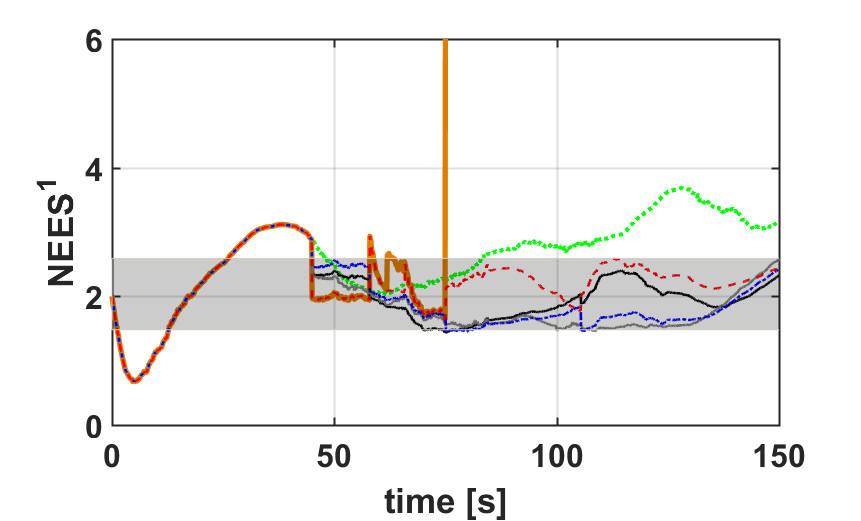}}\\\vspace{-0.1in}
\subfloat[Robot 2]{\includegraphics[scale=0.2]{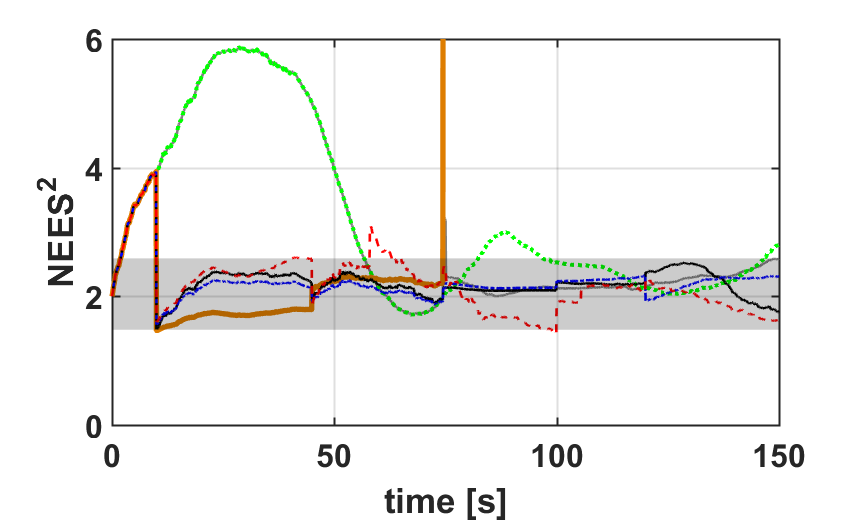}}
\subfloat[Robot 3]{\includegraphics[scale=0.2]{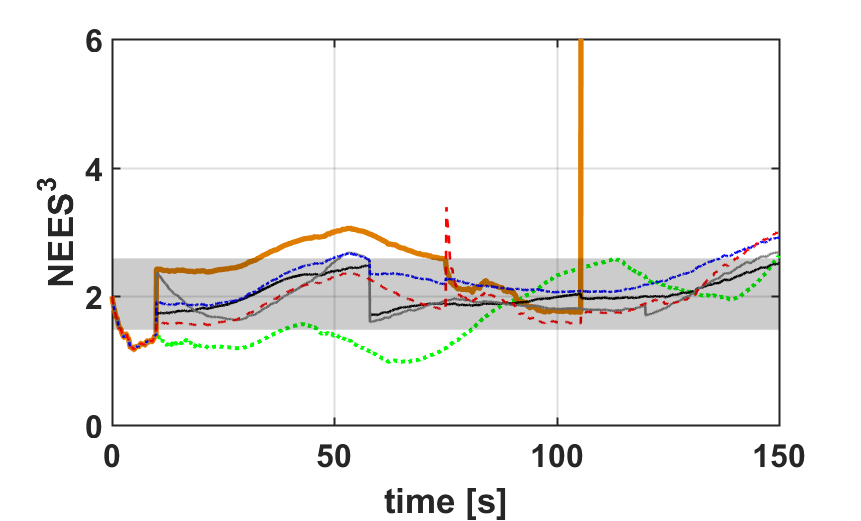}}
\caption{\blue{\small The RSME and NEES plots for our simulation scenario. The shaded area in the NEES plots show the consistency zone. 
}}\vspace{-0.2in}
\label{Fig::error}
\end{figure}

\blue{The simulation results for the position RSME and the NEES plots are shown in Fig.~\ref{Fig::error} for the DMV- (with $\gamma=1$) and PECMV-based CL along with those for the dead-reckoning (DR) only localization, Naive CL in which the relative measurement updates ignore cross-covariances, Joint CL in which the cross-covariances are maintained exactly and the three robots are considered as a joint system,  and finally the CI CL algorithm of~\cite{LCC-EDN-JLG-SIR:13} which uses a CIF-based method. As the RSME plots show CL improves localization accuracy, with the best performance as expected corresponding to the Joint CL. Moreover, as we were expecting, PECMV has a better localization result than DMV. The Naive CL, which ignores the cross-covariances, produces estimates with large errors. From the NEES plots we also see that disregarding prior correlations results in filter inconsistency for the Naive CL. In contrast, the rest of the methods by either exact account of the cross-covariances (Joint CL) or implicit account of the cross-covarainces (DMV, PECMV and CI CL) demonstrate consistent localization results. The RSME and the NEES plots for the DMV-based CL with $\gamma=1-\omega^\star$ has a very similar form as of the CI CL algorithm of~\cite{LCC-EDN-JLG-SIR:13} so they are not shown in Fig.~\ref{Fig::error}. In fact, one can show that the updated covariance matrix of the DMW method with $\gamma=1-\omega^\star$ structurally is very similar to the updated covariance matrix of~\cite{LCC-EDN-JLG-SIR:13}. This can explain why the DMV method with $\gamma=1$ preforms better than the CI CL of~\cite{LCC-EDN-JLG-SIR:13}. }

\blue{The simulation results in Fig.~\ref{Fig::error} show that the PECMV-based CL out-preforms the DMV-based CL in estimation performance. However, this improvement comes with an extra computational cost. Table~\ref{table::execut_time} shows that the average time it takes to run a numerical solver for the optimization problem~\eqref{eq::PECMV-opt-convex} for a sample case from our simulation study is much longer that the one for~\eqref{eq::omega_star}. To solve~\eqref{eq::omega_star} we use the Quadratic Fit algorithm~\cite{DGL:08}, which has a fast superlinear convergence. To solve~\eqref{eq::PECMV-opt-convex} we use the the CVXOPT~\cite{BIBcvxopthome}, a Python software for convex optimization. 
}

\begin{figure}[t]
\centering
\includegraphics[scale=0.26]{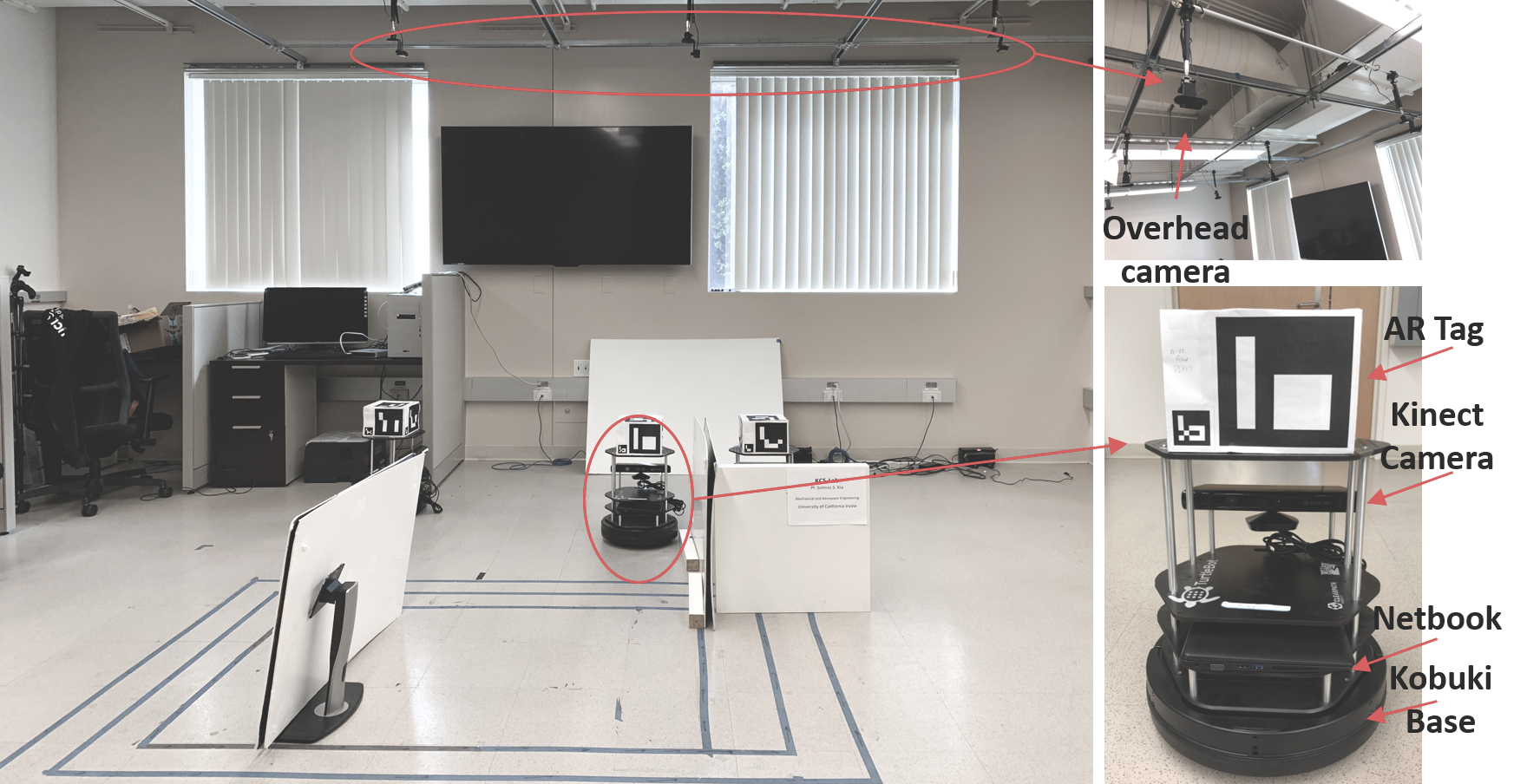}
\caption{\blue{\small The experimental setup: an AR~tags cube atop robots  enables
the Kinect and the overhead cameras to take pose~measurements.}}\vspace{-0.12in}
\label{Fig::testbed}
\end{figure}

\begin{figure}[t]
\centering
\includegraphics[scale=0.45]{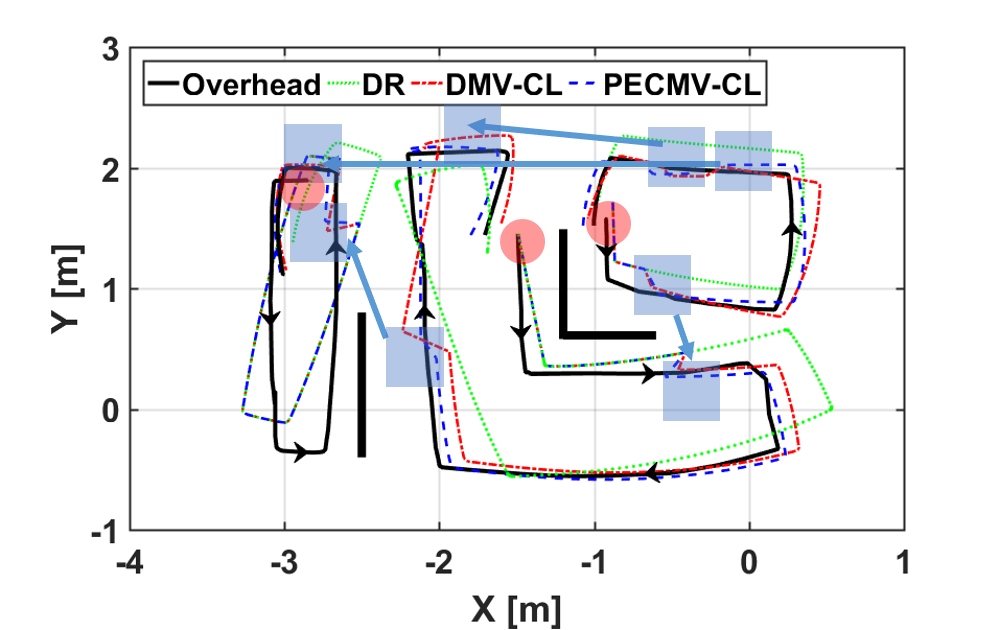}
\caption{\blue{\small
Trajectories of three Turtlebots in our experiment, 
generated by four simultaneously running ROS packages: one for the
overhead camera location tracking (the solid black), one for the dead reckoning (DR) only location estimate (the green dotted curve), and the other two to obtain location estimates by
the the CL via DMV update (the red dash-dotted curve) and the CL via PECMV update (the blue dashed curves). The robots start from a point in the red circles and move in the direction of the black arrows. The relative measurement updates happen at the blue squares and the blue arrows indicate which robot has taken the relative measurement from what other robot.
}}\vspace{-0.12in}
\label{Fig::experimental}
\end{figure}

\blue{
\emph{Experiment}:
We carry out a experimental evaluation of our methods, using three Turtlebot robots operated by the Robot Operating System (ROS), see Fig.~\ref{Fig::testbed}. Each Turtlebot takes self-motion measurement via its onboard wheel encoders with an accuracy level of $\sigma_{v}=\frac{\sqrt{2}}{2}\sigma_{e}V$ and $\sigma_{\omega}=\frac{\sqrt{2}}{a}\sigma_{e}V$ where $a$ is the radius of the wheels, $\sigma_{e}=0.8$ is the accuracy of the wheel-encoder measurements and $V$ is the linear velocity of the robot. Each Robot has a cube of Augmented Reality (AR)~tags atop to enable
the Kinect and the overhead cameras to take pose~measurements. We use the  overhead cameras to track the robots and obtain reference trajectories to assess the localization accuracy of the DMV and the PECMV update~methods. Our vision systems use ROS's ArUco library~\cite{RMS-BM:00} to collect measurement. The accuracy of relative pose measurement based on computer vision is $0.03m$ for position and $6$ degrees for orientation. During our experiment, the robots move along pre-specified trajectories in the active vision zone of the overhead camera system, which is a $4m\times5m$ area. Each Turtlebot collects odometry measurements to dead reckon, and takes a relative pose measurement from other robots when they come to its measurement view. Fig.~\ref{Fig::experimental} shows the outcome of one of our experiments. As seen, the relative measurement updates improve the localization accuracy of the robots, with the PECMV update method preforming better than the DMV update method. A video of this experiment is available at~\cite{Demo_Youtube}.
}


\begin{table}[t]
    \centering
        \caption{\blue{\small Execution time of the numerical solvers of  optimization problems~\eqref{eq::omega_star} (for DMV method with $\gamma=1$) and~\eqref{eq::PECMV-opt-convex} (for the PECMV method) over different computing~platforms.}\vspace{-0.09in}}\label{table::execut_time}
 \begin{tabular}{|p{60mm}||c||c|}
    \hline
    &\multicolumn{2}{c|}{Run time (msec)}\\
    \hline
    Computing~platform&DMV&ECMV\\
    \hline
   Turtlebot netbook: 
    Intel Pentium CPU 2117U@ 1.80GHz, dual-core, 4GHZ memory&3.981
&652.314
\\
    \hline
     MSI GS60 laptop: Intel $\text{Core}^{TM}$ CPU i7-6700HQ@2.60GHz, quad-core, 16GHZ memory&3.902
&612.548\\
    \hline
    Dell Desktop:   
    Intel $\text{Core}^{TM}$ CPU i3-4150@ 3.50GHz, quad-core, 8GHZ memory&1.865&322.104\\
\hline
\end{tabular}

\end{table}

\section{Conclusion}
\vspace{-0.05in}
We considered the problem of cooperative localization for a group of communicating mobile agents, which because of challenging conditions cannot hold any form of network-wide connectivity to maintain an explicit account of their past state estimate correlations. We proposed two relative measurement update methods, which account for past correlations implicitly to ensure the consistency of the localization filter.
In our first solution, we accounted for unknown inter-agent correlations via a well-established upper bound on the joint covariance matrix of the agents. In the second method, we used an optimization framework to estimate the unknown inter-agent cross-covariance matrix. We showed that our second method out-preforms the first one, however, this comes with a higher computational cost. Therefore, the choice between these two methods is a trade of between performance and computational cost. We used our update methods to propose a cooperative localization scheme in which each agent localizes itself in a global coordinate frame using a local filter driven by local dead-reckoning and occasional absolute measurements, and opportunistically corrects its pose estimate whenever a relative measurement takes place between this agent and another mobile agent. In our framework, to process any relative measurement, only the two agents involved in that measurement need to communicate. \blue{Moreover, every agent maintains only its own state estimate, and the relative measurement update process is independent of the size of the network}.

\bibliographystyle{ieeetr}%
\bibliography{library} 

\end{document}